\pdfoutput=1
\documentclass[11pt]{article}

\usepackage[]{acl}

\usepackage{times}
\usepackage{latexsym}

\usepackage[T1]{fontenc}
\usepackage[utf8]{inputenc}

\usepackage{microtype}

\usepackage{inconsolata}

\usepackage{graphicx}
\usepackage[nolist]{acronym}
\usepackage{multirow}

\begin{acronym}
    \acro{rq}[RQ]{research question}
    \acroplural{rq}[RQs]{research questions}
    \acro{nlp}[NLP]{natural language processing}
    \acro{llm}[LLM]{large language model}
    \acro{lora}[LoRA]{low-rank adaptation}
    \acroplural{llm}[LLMs]{large language models}
    \acroplural{plm}[PLMs]{pre-trained language models}
    \acro{ai}[AI]{artificial intelligence}
    \acro{qa}[QA]{Question answering}
    \acro{kbqa}[KBQA]{question answering over knowledge bases}
    \acro{nlg}[NLG]{Natural language generation}
    \acro{kg}[KG]{knowledge graph}
    \acroplural{kg}[KGs]{knowledge graphs}
\end{acronym}

\title{A Comparative Analysis of Conversational Large Language Models in Knowledge-Based Text Generation}

\author{Phillip Schneider$^1$, Manuel Klettner$^1$, Elena Simperl$^2$, and Florian Matthes$^1$ \\
         $^1$Technical University of Munich, Department of Computer Science, Germany \\
         $^2$King’s College London, Department of Informatics, United Kingdom \\
         \texttt{\{phillip.schneider, manuel.klettner, matthes\}@tum.de} \\
         \texttt{elena.simperl@kcl.ac.uk}}

\begin{document}
\maketitle
\begin{abstract}
Generating natural language text from graph-structured data is essential for conversational information seeking. Semantic triples derived from knowledge graphs can serve as a valuable source for grounding responses from conversational agents by providing a factual basis for the information they communicate. This is especially relevant in the context of large language models, which offer great potential for conversational interaction but are prone to hallucinating, omitting, or producing conflicting information. In this study, we conduct an empirical analysis of conversational large language models in generating natural language text from semantic triples. We compare four large language models of varying sizes with different prompting techniques. Through a series of benchmark experiments on the WebNLG dataset, we analyze the models' performance and identify the most common issues in the generated predictions. Our findings show that the capabilities of large language models in triple verbalization can be significantly improved through few-shot prompting, post-processing, and efficient fine-tuning techniques, particularly for smaller models that exhibit lower zero-shot performance.
\end{abstract}

\section{Introduction}
Accessing structured information through natural language interfaces has garnered significant research interest in \ac{nlp} \cite{aliannejadi2021analysing,radlinski2017theoretical}. 
For instance, the emerging information retrieval paradigm of conversational search frames information-seeking processes within multi-turn dialogue interactions. 
Conversational search facilitates exploring and progressively narrowing the search scope to relevant knowledge items within an information space. 
These search-oriented conversational interfaces are often connected to structured data sources like knowledge graphs. However, a key challenge lies in mediating between natural language, in which users express their queries, and machine-readable knowledge representations. The task of data-to-text generation focuses on this issue, taking structured data as input to produce coherent, human-readable text, which has been extensively studied with approaches ranging from rule-based to supervised neural network-based techniques.

Over the last years, the field of \ac{nlp} has witnessed a shift in methodologies with the advent of pre-trained \acp{llm}. Unlike traditional supervised learning approaches that rely on annotated datasets, \acp{llm} are trained in a self-supervised manner, predicting tokens within vast amounts of unlabeled data. Combined with scaling up the model size and training corpora, this approach has demonstrated remarkable emergent capabilities of \acp{llm} and their prowess in multi-task learning \cite{radford2019language,brown2020language}. An advantage of \acp{llm} lies in prompt-based (in-context) learning. Through carefully defined prompts, these foundation models can perform multiple tasks like question-answering, semantic parsing, or text summarization \cite{pengfei2023}. More recently, there has been a growing interest in optimizing \acp{llm} for conversational interactions by pre-training on dialogue corpora, instruction fine-tuning, and reinforcement learning from human feedback \cite{thoppilan2022lamda,openai2022chat}. 

Although \acp{llm} offer tremendous potential for conversational interaction, owing to their ability to produce responses for arbitrary input, they have known limitations, such as the risk of hallucinating or omitting important information and a lack of transparency regarding the origins of information sources from which they derive their outputs \cite{dou-etal-2022-gpt,ji2023survey}. In order to mitigate these limitations, it becomes imperative to ground their generated outputs in verifiable factual data from knowledge graphs. However, there has been insufficient systematic investigation into their proficiency in verbalizing graph-structured data input.

To assess \acp{llm} in knowledge-based text generation, we compare four models of different sizes and training objectives, with a primary focus on models optimized for conversational interaction. Based on the popular WebNLG benchmark dataset, we evaluate the models' performance in generating natural language text from semantic triples. Through multiple experiments, we analyze different configurations of models and prompting techniques, discussing insights about their individual capabilities and limitations. Our contributions include: (1) adapting the WebNLG benchmark to evaluate closed- and open-source \acp{llm}, (2) providing a thorough error analysis and insights on model performance with automatic reference-based metrics as well as human evaluation, and (3) creating a new fine-tuning dataset with 26,422 conversations with triple-to-text verbalizations in chat completion format. To ensure reproducibility, we publish our source code and datasets in a GitHub repository.\footnote{\href{https://github.com/sebischair/LLM-KG-D2T}{GitHub: https://github.com/sebischair/LLM-KG-D2T}}

\section{Related Work}
Existing works from the \ac{nlp} literature have explored knowledge-based text generation, with significant advancements driven by new deep learning architectures and fine-tuning language models on downstream tasks \cite{li2021survey,schneider-etal-2022-decade}. For triple-to-text generation, many evaluations use the established WebNLG benchmark \cite{colin-etal-2016-webnlg}. Several studies have focused on comparing neural pipeline versus end-to-end approaches, assessing supervised versus unsupervised training regimes, and developing frameworks for making text generation more controllable through neuro-symbolic methods \cite{castro-ferreira-etal-2019-neural,schmitt-etal-2020-unsupervised,moryossef-etal-2019-step,su-etal-2021-plan-generate}.

Concerning pre-trained language models, \citet{chen-etal-2020-shot} were among the first to propose the task of few-shot natural language generation. With just 200 table-to-text training examples, their approach achieves strong performance and good generalization. By collecting a novel dataset and experimenting with few-shot fine-tuning, \citet{kasner-etal-2023-mind} demonstrate that pre-trained language models trained with a diverse set of labels exhibit robustness in verbalizing knowledge graph relations, being capable of generalizing to novel domains. 
Another study from \citet{liu-etal-2021-awakening} highlights the ability of \acp{plm} to uncover hidden mappings between linguistic tokens and real-world concepts. Conducting experiments on four datasets, the authors show the effectiveness of their awakening latent grounding approach for generating structured queries from text. 
Similar to our work, \citet{han2023pive} assess capabilities of \acp{llm} but for text-to-graph generation with the GPT-3.5-Turbo model. They develop a prompting framework with iterative verification, improving the generation quality. In contrast, our objective is to achieve a comprehensive understanding of conversational \acp{llm} for triple verbalization rather than solely concentrating on individual use cases or models. To the best of our knowledge, we are the first to conduct a comparative analysis of conversational \acp{llm} and prompt configurations on the task of triple-to-text generation. The empirical approach employed in this study is related to our previous work on evaluating \acp{llm} for semantic parsing for conversational question answering over knowledge graphs \cite{schneider-etal-2024-evaluating}.

\begin{table*}[t!]
\small
\centering
\begin{tabular}{l@{}ccccc@{}ccccc}
\hline
\textbf{Model} & & \multicolumn{4}{c}{\textbf{Zero-Shot Prompt}} & & \multicolumn{4}{c}{\textbf{Few-Shot Prompt}} \\
\cline{3-6}
\cline{8-11}
 & & BLEU & METEOR & TER & BERTScore & & BLEU & METEOR & TER & BERTScore \\
\hline
LLaMA-7B & & 0.06 & 0.21 & 1.03 & 0.84 & & 0.11 & 0.26 & 1.03 & 0.85 \\
LLaMA-7B + PP & & 0.15 & 0.25 & 0.76 & 0.89  & & 0.38 & 0.36 & 0.53 & 0.94  \\
Vicuna-7B & & 0.27 & 0.35 & 0.68 & 0.92  & & 0.39 & 0.38 & 0.64 & 0.93  \\
Vicuna-7B + PP & & 0.27 & 0.35 & 0.68 & 0.92  & & 0.43 & 0.39 & 0.51 & 0.95  \\
LLaMA-FT-7B & & 0.47 & 0.40 & 0.55 & 0.94  & & 0.47 & 0.40 & 0.55 & 0.94  \\
LLaMA-FT-7B + PP & & \textbf{0.52} & \textbf{0.41} & \textbf{0.42} & \textbf{0.96}  & & \textbf{0.53} & \textbf{0.41} & \textbf{0.42} & \textbf{0.96}  \\
GPT-3.5-Turbo & & 0.41 & \textbf{0.41} & 0.56 & 0.95  & & 0.39 & 0.40 & 0.65 & 0.94  \\
GPT-3.5-Turbo + PP & & 0.41 & \textbf{0.41} & 0.56 & 0.95  & & 0.44 & \textbf{0.41} & 0.50 & 0.95  \\
\hline
Copy-Baseline & & 0.02 & 0.02 & 0.95 & 0.79  & & 0.02 & 0.02 & 0.95 & 0.79  \\

\hline
\end{tabular}
\caption{Zero-shot and few-shot performance metrics on WebNLG test set evaluated by BLEU, METEOR, TER, and BERTScore-F1 (+ PP denotes post-processed model output). Bold values indicate the best value per metric.}
\label{tab:d2t-results}
\end{table*}

\section{Experiments}
\paragraph{Experimental Setup}
We conduct our experiments on the \textit{WebNLG+ 2020} dataset, a DBpedia-based triple-to-text benchmark with a total of 1,779 test examples \cite{castro-ferreira-etal-2020-2020}. As evaluation metrics, we calculate the lexical similarity between model outputs and human annotations using \textit{BLEU} \cite{papineni-etal-2002-bleu}, \textit{METEOR} \cite{banerjee-lavie-2005-meteor}, and \textit{TER} \cite{snover-etal-2006-study}. Since these metrics mainly focus on lexical overlaps, we also use the \textit{BERTScore} metric, which captures semantic similarity \cite{zhang2020BERTScore}.

As a commercial state-of-the-art \ac{llm}, we include \textit{GPT-3.5-Turbo} (\textit{ChatGPT}) \cite{openai2022chat} in our comparison. It is optimized for conversations and has demonstrated remarkable zero-shot performance on various \ac{nlp} tasks. Consequently, it is often used as a benchmark for comparing \acp{llm}. 
We ran our experiments with the model released in June 2023 (GPT-3.5-Turbo-0613). Further, we opted to test \textit{LLaMA}, a collection of open-source \acp{llm} from Meta \cite{touvron2023llama}, achieving competitive performance on various benchmarks. We include three model variations with 7B parameters of the first LLaMA version. In addition to the non-conversational base model (\textit{LLaMA-7B}), we included a fine-tuned model (\textit{LLaMA-FT-7B}) which we trained on WebNLG examples in a conversational format. To have a sufficiently large fine-tuning corpus, we created a new dataset encompassing 26,422 conversations from all 13,211 WebNLG training examples. We ensured that each triple-to-text example appeared, on average, five times in different contexts. The conversations have different lengths and contain verbalizations from various triple categories. The training was done through \textit{\ac{lora}}, a method that fine-tunes only a subset of the model's parameters, referred to as low-rank matrices, rather than updating the entire parameter space, improving the fine-tuning efficiency \cite{hu2022lora}. During training time, the model takes in a full conversation in chat completion format, characterized by a series of turns attributed to the user or assistant role (i.e., the model learns from a sequence of sequence-to-sequence examples). We employed five training epochs, a per-device training batch size of eight, and used a half-precision floating-point format (FP16). Another fine-tuned LLaMA model we compared is \textit{Vicuna}. It was trained on a corpus of around 70K user-shared ChatGPT conversations crawled from the ShareGPT website. 
Preliminary evaluations from \citet{chiang2023vicuna} demonstrate that Vicuna exhibits a higher level of detail and structure in its responses than LLaMA, highlighting the advantage of fine-tuning on dialogue data.

The LLaMA and Vicuna models are prompted in the chat completion structure of the FastChat\footnote{\href{https://github.com/lm-sys/FastChat}{FastChat: https://github.com/lm-sys/FastChat}} platform, replicating OpenAI's chat completion API endpoint with a structured list of system, user, and assistant messages. We set the token limit to 128 and the temperature parameter to 0, maximizing deterministic generation by favoring high-probability words. The zero-shot prompt contains only the following system message with a triple verbalization instruction: \textit{``SYSTEM: Generate a concise text for the given set of triples. Ensure that the generated output only includes the provided information from the triples.''}. 
The few-shot prompt expands the instruction with three in-context examples provided as user and assistant messages in the format: \textit{``USER: Input triples: {[}\{'object': 'Mike\_Mularkey','property': 'coach','subject': 'Tennessee\_Titans'\}{]} 
\newline
\textit{``ASSISTANT: Output text: Mike Mularkey is the coach of the Tennessee Titans.''}}
Table~\ref{tab:prompts} in Appendix~\ref{sec:appendix-a} displays each prompt in full length.

\paragraph{Results of Performance Metrics}
Table~\ref{tab:d2t-results} summarizes the calculated metrics. The Copy-Baseline denotes copying the triples as output without processing. It is included as a metric reference point to establish a lower bound \cite{kasner-dusek-2022-neural}. We distinguish between scores for raw and post-processed (+ PP) outputs. Post-processing involved the removal of ``Output text'' or ``Output'' since they are not intended parts of the desired text prediction but were present in the few-shot prompt. Additionally, repeated instructions or in-context examples from the prompt were removed when they appeared in the generated output.

Examining the scores, LLaMA-FT-7B demonstrates superior performance compared to the other models. Even without few-shot examples, it effectively learned from fine-tuning to handle the triple verbalization task, gaining only a minor performance increase through few-shot prompting. The second-ranking model, GPT-3.5-Turbo, shows similar scores, which is remarkable because it was not explicitly trained for triple-to-text generation. Notably, Vicuna achieves a performance level almost on par with the much bigger GPT-3.5-Turbo model when it was provided with in-context examples and the output was post-processed. In the zero-shot setting, Vicuna could not match the scores of GPT-3.5-Turbo but outperformed LLaMA-7B. Although LLaMA is the worst-performing model, it claims the most significant improvements through few-shot prompting and post-processing, with scores not too far from Vicuna. The metrics collectively suggest that all tested \acp{llm} can generate reasonable output text from knowledge graph triples. Besides, we observe that while all models show improvements with few-shot prompting or post-processing, models trained on conversations like Vicuna require less post-processing and exhibit better zero-shot proficiency, resulting in comparatively smaller performance gains from post-processed outputs or in-context examples.

\paragraph{Analysis and Discussion}
Our experiments reveal that \acp{llm}, especially those fine-tuned on conversations, are capable of triple-to-text generation without explicit training. However, as expected, the fine-tuned LLaMA-FT-7B model achieved the best overall performance. 
The WebNLG triple verbalization task involves different subtasks, such as segmentation of the input data, lexicalization of the DBpedia properties, information aggregation, and surface realization of grammatically correct text \cite{colin-etal-2016-webnlg}. All of these subtasks are handled by \acp{llm} in an end-to-end manner. In direct comparison to state-of-the-art models evaluated on WebNLG like \textit{Control Prefixes} (BLEU: 0.62, METEOR: 0.45, TER: 0.35) from \citet{clive-etal-2022-control} or \textit{T5-Large+Wiki+Position} (BLEU: 0.61, METEOR: 0.44, TER: 0.36, BERTScore: 0.96) from \citet{wang-etal-2021-stage}, the \acp{llm}' lexical similarity metrics are worse. Yet, when looking at semantic similarity, the BERTScore metric of the LLaMA-FT-7B model is identical at 0.96. We hypothesize that the lower lexical similarity is partly caused by the concise writing style of the WebNLG human ground truth verbalizations, aggregating as much information as possible in succinct sentences. While many WebNLG annotations are as short as possible (e.g., \textit{``The 98.0 minute film Super Capers starring Danielle Harris was written by the director Ray Griggs.''}), the more verbose output of \acp{llm} like GPT-3.5-Turbo consists of multiple sentences (e.g., \textit{``Danielle Harris stars in the movie Super Capers. The writer of the movie is Ray Griggs. The movie has a runtime of 98.0 minutes.''}). This concise writing style can be better learned and replicated by LLaMA-FT and other fine-tuned models. We also observed that the \acp{llm} had a tendency to occasionally use passive voice, initiating sentences with the object because the input triples were ordered as (\textit{object, property, subject}), whereas the human annotators started with the subject using an active voice structure. This might be another factor of lower lexical similarity metrics, although the semantic content was the same.

\begin{figure}[t]
  \centering
  \includegraphics[width=0.93\linewidth]{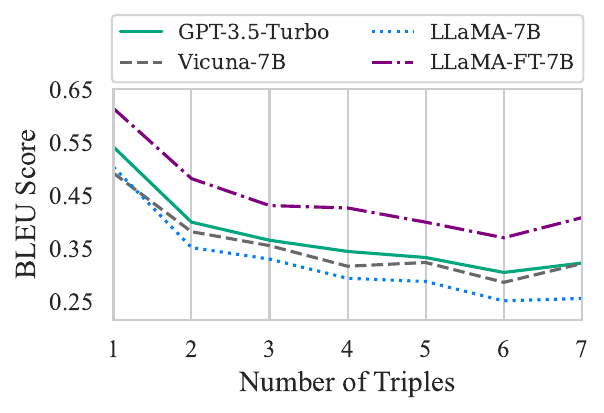}
  \caption{Comparison of BLEU score by number of triples for few-shot models with post-processing.}
\label{fig:bleu-line}
\end{figure}

With a larger number of input triples, models struggle more to transform structured information into cohesive text. Figure~\ref{fig:bleu-line} illustrates the decreasing model performance when confronted with multiple triples. While all four \acp{llm} follow the same trend, the performance loss seems to be a tapering decrease. Besides, we analyzed model performance differences across the 16 triple categories and found a similar pattern that the worst-performing categories, such as \textit{Food}, \textit{SportsTeam}, or \textit{ComicsCharacter} also had the highest average triple count per example. Since aggregating information into short sentences is also desired in conversational user interactions, we compared the sentence count of generated predictions for each model regarding the number of input triples. As can be discerned from Figure~\ref{fig:sentences-scatter} in Appendix~\ref{sec:appendix-a}, the fine-tuned LLaMA-FT model produces sentences in direct proportion to the number of input triples in alignment with the human annotations. Vicuna and GPT-3.5-Turbo, which have been explicitly trained on conversation data, exhibit similar generation behavior. While LLaMA-FT produces the fewest sentences, Vicuna seems to be a bit less verbose than GPT-3.5-Turbo. In contrast, text outputs from LLaMA contain, on average, the largest number of sentences and show a much higher variance. This suggests that fine-tuning \acp{llm} on instructions from dialogue corpora improves adherence to concise triple verbalization.

\begin{table*}[h]
\centering
\begin{tabular}{lcccc}
\hline
\textbf{Issue Type} & \textbf{LLaMA-7B} & \textbf{Vicuna-7B} & \textbf{LLaMA-FT-7B} & \textbf{GPT-3.5-Turbo}\\
 & \multicolumn{3}{c}{relative frequency: zero-shot / few-shot} \\
\hline
Inaccurate & 0.60* / 0.61 & 0.41* / 0.48 & 0.19 / 0.17 & 0.13 / 0.11 \\
Mistranslated & - / - & 0.01* / - &  - / - & - / - \\
Off-prompt & 0.65 / - & 0.27 / - & - / - & - / - \\
Redundant & 0.23* / 0.07 & 0.02* / - & - / 0.01 & 0.01 / 0.01 \\
Unlexicalized & 0.69* / - & 0.27* / - & - / - & 0.07 / - \\
\hline
\end{tabular}
\caption{Relative frequency of issue types for zero-shot and few-shot prompts in evaluated sample of 150 predictions with lowest averaged BLEU and METEOR scores. For values marked with ``*'', the relative frequency only considers generations being on-prompt.}
\label{tab:issue-distribution}
\end{table*}

After conducting the automatic evaluation, we manually examined the model predictions to gauge their reliability and grouped the most common issues into five types as presented in Table~\ref{tab:issue-types} in Appendix~\ref{sec:appendix-a}. For example, the \acp{llm} sometimes misinterpreted the prompt, failed to lexicalize triples correctly, or produced inaccurate information. Most of these issues occurred in zero-shot predictions from LLaMA or Vicuna, whereas GPT-3.5-Turbo produced the most reliable outputs. To obtain more profound insights into the model-specific occurrence rates of the issue types, two researchers jointly evaluated a sample of 75 zero- and 75 few-shot predictions for the lowest averaged BLEU and METEOR scores across all models. The obtained results are summarized in Table~\ref{tab:issue-distribution}. Looking at the relative frequencies, it can be seen that the LLaMA base model has the highest incidence of issues from all types, followed by Vicuna and then LLaMA-FT with better reliability, and GPT-3.5-Turbo as the most dependable model.

As to be expected from instruction-tuned and fine-tuned models, LLaMA-FT, Vicuna, and GPT-3.5-Turbo demonstrate a much greater ability to generate zero-shot output that aligns with the given prompt. Conversely, LLaMA tended to misinterpret the prompt, failing to produce the desired output format in nearly two-thirds of the evaluated instances (0.65). Interestingly, off-prompt issues could be effectively resolved in all models by including few-shot examples in the prompt. While few-shot prompting reduced off-prompt generations and caused the \acp{llm} to produce actual sentences based on the graph triples, this led to a relative increase of inaccurate generations, such as hallucinated information, twisted numbers, or omitted facts from the triples. Occasionally, the relationships within these triples were also compromised. The rate of inaccurate zero-shot output in LLaMA (0.60) and Vicuna (0.41) was three to four times higher in comparison to GPT-3.5-Turbo (0.13). 

Another issue type where the usefulness of few-shot examples became evident is unlexicalized triples, meaning the translation of entities and relations into their intact word form. This was observed across all models except LLaMA-FT, with LLaMA and Vicuna particularly affected. Providing in-context examples with lexicalized triples could completely resolve unlexicalized triples for all models. Problems with redundancy, which involves the unnecessary repetition of information, are mostly associated with LLaMA. This was due to some instances where LLaMA became stuck in a loop, repeatedly generating the same sequence until the maximum token limit was reached. In contrast, this issue type appears to be less of a problem for the other models. Lastly, there are rare cases in which the \ac{llm} generated output in a language other than the prompt language English. This happened, for example, when most of the input triples contained words in Spanish. Only Vicuna faced translation issues in our benchmark test, specifically in zero-shot scenarios. This behavior may be attributed to its diverse fine-tuning dataset that contains text translation instructions.

\section{Conclusion }
We compared the abilities of \acp{llm} in knowledge-based text generation. Our results indicate that even smaller 7B-\acp{llm} exhibit reasonable performance in verbalizing triples, conveying intended meanings and facts in a coherent manner, although they might not always be factually accurate or perfectly replicate the writing style of human annotations. We also discussed model-specific differences and common generation issues that can be mitigated through few-shot prompting or fine-tuning. In future work, we plan to investigate how our findings generalize to more complex graph data structures. 

\section{Limitations}
Our comparative analysis has certain limitations. We focus solely on text generation based on knowledge graph triples, and we acknowledge that verbalizing entire subgraphs or producing graph queries are other important tasks worth exploring. Nonetheless, by studying semantic triples, we can still derive valuable insights about the performance of \acp{llm} for processing more complex graph data structures. In that regard, it is recommended to expand the comparison with human evaluations that go beyond automatically calculated metrics and to assess more models, particularly those trained on source code or documents with structured data.

Further, the employed test dataset is limited to English triples. Since pre-training corpora of \acp{llm} primarily consist of English text data, they likely work better where entities and relations correspond to meaningful English words or morphemes. Consequently, it is to be expected that \acp{llm} exhibit worse performance on multilingual benchmarks with more morphologically rich languages, such as Russian, which is also part of the WebNLG dataset.

\section{Ethical Considerations}
Our experiments were conducted on the publicly available WebNLG dataset, ensuring that no demographic or identifying information about individuals was processed or disclosed. Because our focus was not on addressing well-documented issues like privacy or biases associated with \acp{llm}, we acknowledge potential risks and concerns in line with similar studies dealing with \acp{llm}. The experiments with LLaMA, LLaMA-FT, and Vicuna were executed on a single NVIDIA V100 GPU and required relatively low computational resources, with around one GPU hour of inference time per model.   

\section*{Acknowledgements}
This work has been supported by the German Federal Ministry of Education and Research (BMBF) Software Campus grant 01IS17049.

\bibliography{anthology,custom}

\newpage
\onecolumn
\appendix
\section{Appendix}
\label{sec:appendix-a}
\begin{figure*}[h]
  \centering
  \includegraphics[width=0.99\linewidth]{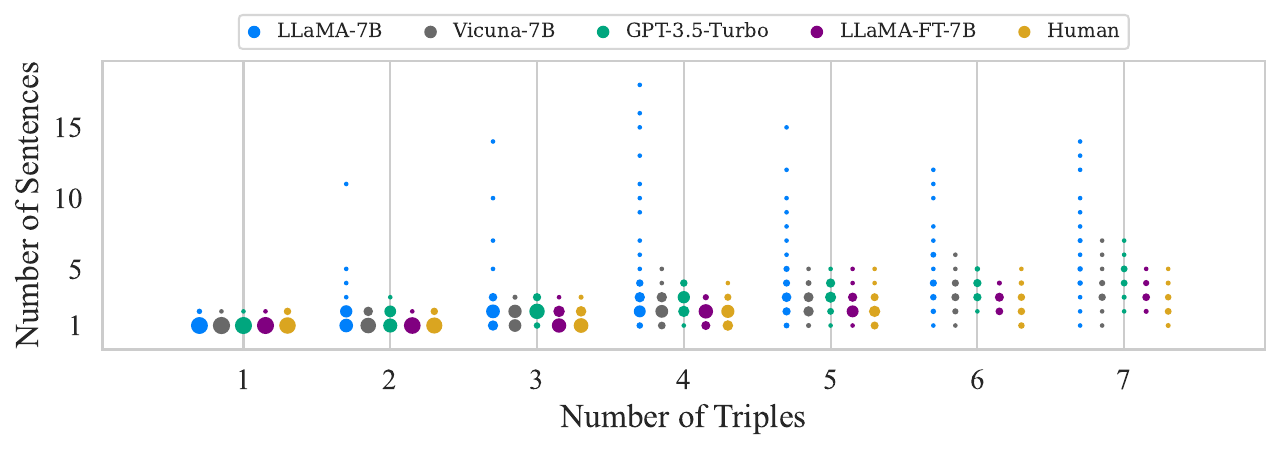}
  \caption{Distribution of model generated sentences by number of triples for few-shot models with post-processing. The size of the dots reflects the occurrence frequency. The ground truth annotations are denoted as ``Human''.}
\label{fig:sentences-scatter}
\end{figure*}

The Appendix provides further insights into the results of our research, including the model prompts in full length (Table~\ref{tab:prompts}), an overview of common issue types identified in the predictions (Table~\ref{tab:issue-types}), and a comparative distribution chart of generated sentences by number of triples for each model (Figure~\ref{fig:sentences-scatter}).

\begin{table*}[h]
\centering
\begin{tabular}{lp{13cm}}
\hline
\textbf{Prompt Type} & \textbf{Prompt Content}\\
\hline
Zero-shot & \verb|SYSTEM:| Generate a concise text for the given set of triples. Ensure that the generated output only includes the provided information from the triples.
\\
\hline
Few-shot & \verb|SYSTEM:| Generate a concise text for the given set of triples. Ensure that the generated output only includes the provided information from the triples. 
\newline \newline
\verb|USER:| Input triples: {[}\{'object': 'Mike\_Mularkey','property': 'coach','subject': 'Tennessee\_Titans'\}{]}
\newline \newline
\verb|ASSISTANT:| Output text: Mike Mularkey is the coach of the Tennessee Titans.
\newline \newline
\verb|USER:| Input triples: {[}\{'object': 'Albert\_E.\_Austin', 'property': 'successor', 'subject': 'Alfred\_N.\_Phillips'\}, \{'object': 'Connecticut', 'property': 'birthPlace', 'subject': 'Alfred\_N.\_Phillips'\}, \{'object': 'United\_States\_House\_of\_Representatives', 'property': 'office', 'subject': 'Alfred\_N.\_Phillips'\}{]}
\newline \newline
\verb|ASSISTANT:| Output text: Albert E. Austin succeeded Alfred N. Phillips who was born in Connecticut and worked at the United States House of Representatives.
\newline \newline
\verb|USER:| Input triples: {[}\{'object': 'College\_of\_William\_\&\_Mary', 'property': 'owner', 'subject': 'Alan\_B.\_Miller\_Hall'\}, \{'object': '2009-06-01', 'property': 'completionDate', 'subject': 'Alan\_B.\_Miller\_Hall'\}, \{'object': '101 Ukrop Way', 'property': 'address', 'subject': 'Alan\_B.\_Miller\_Hall'\}, \{'object': 'Williamsburg,\_Virginia', 'property': 'location', 'subject': 'Alan\_B.\_Miller\_Hall'\}, \{'object': 'Robert\_A.\_M.\_Stern', 'property': 'architect', 'subject': 'Alan\_B.\_Miller\_Hall'\}{]}
\newline \newline
\verb|ASSISTANT:| Output text: The Alan B Miller Hall's location is 101 Ukrop Way, Williamsburg, Virginia. It was designed by Robert A.M. Stern and was completed on 1 June 2009. Its owner is the College of William and Mary.
\\
\hline
\end{tabular}
\caption{Overview of applied zero-shot and few-shot prompts.}
\label{tab:prompts}
\end{table*}

\begin{table*}[h]
\centering
\begin{tabular}{p{1.9cm}p{3cm}p{9.5cm}}
\hline
\textbf{Issue Type} & \textbf{Definition} & \textbf{Example}\\
\hline
Inaccurate & Prediction is incomplete or factually incorrect. & \verb|GT:| Binignit is a dessert that comes from the Philippines. \newline
\verb|PRED:| Philippines is a dessert.\\
\hline
Mistranslated & Triples were verbalized in a language other than English. & \verb|GT:| The Mexican based city of Ciudad Ayala in Morelos stands at 1147 metres above sea level and is in the PDT time zone. \newline
\verb|PRED:| Ciudad Ayala es parte de Morelos, que es un estado en México. Ciudad Ayala tiene una elevación de 1147 metros sobre el nivel del mar [...] \\
\hline
Off-prompt & Prediction is unrelated to the prompt and contradicts desired output format.  & \verb|GT:| GMA Network is a public company. \newline
\verb|PRED:| The input triples are: {[}\{'object': 'Public\_company', 'property': 'type', 'subject': 'GMA\_Network\_(company)'\}{]}. \\
\hline
Redundant & Unnecessary repetition of triple information. & \verb|GT:| The Pontiac Rageous, a car with a coupe body style, assembled in Detroit Michigan, was first and last produced in 1997.
\newline \verb|PRED:| 1997 Pontiac Rageous was produced in Detroit, Michigan, in 1997. It was a coupe. It was assembled in 1997. It was produced in 1997 [...] It was a coupe. It was assembled in 1997 [...]\\
\hline  
Unlexicalized & Entities or relations are not lexicalized. & \verb|GT:| The Fellowship of the Ring was followed by The Two Towers. \newline
\verb|PRED:| The\_Fellowship\_of\_the\_Ring was followed by The\_Two\_Towers. \\
\hline
\end{tabular}
\caption{Overview of five identified issue types with examples from generated model predictions (PRED) and human ground truth annotations (GT).}
\label{tab:issue-types}
\end{table*}

\end{document}